# NONLINEAR MIXTURE MODEL MOTIVATED SUBSPACE CLUSTERING

*Ivica Kopriva*

Division of Computing and Data Science, Ruđer Bošković Instute, Zagreb, Croatia

## ABSTRACT

We derive the linear union-of-subspaces (UoS) model for subspace clustering (SC) from the nonlinear mixture model (NMM) used in blind source separation (BSS) to represent a $D$-dimensional observation vector as an unknown multivariate nonlinear mapping of $C$ latent variables. Assuming the mapping is differentiable up to an unknown order $K$, we approximate NMM by a $K$-th order Taylor expansion, yielding a model equivalent to the linear UoS framework underlying SC. This establishes that: (*i*) the smoothness order $K$ corresponds to the unknown subspace dimension $d$; (ii) $KC$ equals the number of anchors; and (*iii*) the sparsity of the representation vector equals $K$ (i.e., $d$). These relationships enable estimation of bounds on subspace dimension, and that is validated on six benchmark datasets using five established SC algorithms. Established theoretical results are important for post-processing of self-representation matrices estimated by SC algorithms.

## 1. INTRODUCTION

Clustering high-dimensional data into disjoint, homogeneous groups is a fundamental problem in data analysis [1], [2]. It is central to numerous applications, including image segmentation [3] and data mining [4]. In many applications, data exhibit far fewer degrees of freedom than their ambient dimensionality [5], motivating the search for low-dimensional structures embedded in high-dimensional spaces [2]. Accordingly, data points are commonly assumed to lie near a union of multiple low-dimensional subspaces (UoS) [6]–[9]. Exploiting this structure, numerous linear subspace clustering (SC) algorithms have been developed [6]–[12]. These methods first compute pairwise similarities between data samples and then apply spectral clustering to the resulting affinity graph [13]. Ideally, the affinity matrix exhibits a block-diagonal structure in which samples from the same subspace are densely connected, whereas samples from different subspaces are disconnected [9], [14], [15]. In practice, however, a single SC model rarely achieves this ideal structure, making post-processing of the computed affinity matrix desirable [14], [15]. This post-processing step, however, depends on the generally unknown dimension of the underlying low-dimensional subspaces associated with the dataset; see Section 3.2 for details.

These observations motivated us to derive the linear UoS model for SC from a nonlinear generative manifold model, also known as the nonlinear mixture model (NMM) in blind source separation (BSS) [16]. As shown in Section 3.1, an NMM describing $D$ observed mixtures generated from $C$ latent sources can be transformed into a model of $D$ linear mixtures of $\sum_{k=1}^{K} C^k$ statistically dependent sources, where $K$ denotes the order of the Taylor expansion of the observation vector [17]. The parameter $K$ characterizes the smoothness of the unknown nonlinear mapping and is assumed to be small. Theoretical analysis shows that $K$ coincides with the unknown subspace dimension $d$, Section 3.2, while the number of anchors satisfies $M = KC$, Section 3.3. Furthermore, the sparsity constraint imposed on the representation matrix in bipartite graph-based SC is also shown to equal $K$, Section 3.3. Combined with empirical findings reported in [11] [12], [18]–[23], these theoretical results enable estimation of the subspace dimensions for benchmark datasets containing face images (EYaleB [24] and ORL [25]), object images (COIL20 and COIL100 [26]), and handwritten digits (MNIST [27] and USPS [28]). The estimates are validated through $d$-based post-processing [14] of self-representation matrices produced by five SC algorithms [6]–[8], [12] on the six benchmark datasets. The estimated subspace dimensions are also consistent with the empirically determined sparsity ranges reported in [11]. The MATLAB implementation of the proposed method is available at https://github.com/ikopriva/NMMSC.

## 2. BACKGROUND AND RELATED WORK

### 2.1 Subspace clustering

The SC model assumes that data points are drawn from $C \geq 1$ affine subspaces:

$$S_c = \left\{ \mathbf{x}_n \in \mathbb{R}^{D\times 1} : \mathbf{x}_n = \mathbf{A}_c \mathbf{z}_n + \boldsymbol{\mu}_c \right\}_{c=1}^{C} \quad n = 1, ..., N \tag{1}$$

of dimensions $\left\{ d_c < D \right\}_{c=1}^{C}$ [10]. $N$ in (1) denotes the number of data samples, $\mathbf{A}_c \in \mathbb{R}^{D\times d_c}$ denotes a basis of $d_c$-dimensional subspaces, and $\mathbf{z}_n \in \mathbb{R}^{d_c \times 1}$ denotes the representation of data vector $\mathbf{x}_n$ in the subspaces spanned by $\mathbf{A}_c$. When $\{\boldsymbol{\mu}_c = \mathbf{0}\}_{c=1}^{C}$ subspaces are linear. As shown in [29], however, as it is shown in [29], the affine constraint has negligible effect on clustering when $D \ll \sum_{c=1}^{C} d_c$; therefore, a linear model can be assumed. Assuming a self-expressive linear model with Gaussian error term many SC algorithms are be formulated as:

$$\min_{\mathbf{C}} \tfrac{1}{2}\|\mathbf{X}-\mathbf{XC}\|_F^2 + \lambda f\left(\boldsymbol{\sigma}(\mathbf{C})\right) + \tau g(\mathbf{C}) \text{ s.t. } \operatorname{diag}(\mathbf{C}) = \mathbf{0}. \quad (2)$$

In (2), $\mathbf{X} \in \mathbb{R}^{D\times N}$ denotes a dataset of $N$ samples in $D$-dimensional ambient space, $\mathbf{C} \in \mathbb{R}^{N\times N}$ is the self-representation matrix with respect to $\mathbf{X}$, and $\boldsymbol{\sigma}(\mathbf{C})$ denotes the vector of singular values of $\mathbf{C}$. The functions $f$ and $g$ impose low-rank and sparsity constraints on $\mathbf{C}$, respectively. The constraint diag($\mathbf{C}$)=0 prevents each data point to be represented by itself. For the low-rank representation (LRR) SC [6], and the LRR $S_{1/2}$ and $S_{2/3}$ SC algorithms [12] $\tau$=0, while $\ell_1$-, $\ell_{1/2}$- and $\ell_{2/3}$-norm constrains are imposed on $\boldsymbol{\sigma}(\mathbf{C})$. For sparse SC (SSC) [7] $g(\mathbf{C}) = \|\mathbf{C}\|_1$ and $\lambda$=0. For the generalization of minimax concave penalty low-rank sparse SC (GMC-LRSSC) [8] $\lambda+\tau$=1 and both $f$ and $g$ impose constraints on $\boldsymbol{\sigma}(\mathbf{C})$ and $\mathbf{C}$ based on GMC penalty function [30]. Once $\mathbf{C}$ has been estimated the affinity matrix is computed as:

$$\mathbf{W} = \left(|\mathbf{C}| + |\mathbf{C}|^{\mathrm{T}}\right)/2. \quad (3)$$

Spectral clustering [13] is then applied to the graph Laplacian constructed from $\mathbf{W}$ to assign labels to data.

### 2.2 Post-processing of representation matrix

Equation (2) shows that different SC algorithms estimate representation matrix $\mathbf{C}$ from the data matrix $\mathbf{X}$. Ideally, the coefficient $c_{ij}$ should be large when $\mathbf{x}_i$ and $\mathbf{x}_j$ belong to the same subspace (cluster) and zero otherwise. In practice, however, the estimated representation matrix deviates from this ideal one due to model mismatch, noise, and estimation errors. Consequently, post-processing of the learned representation matrix is often employed to improve clustering performance. That is why post-processing of $\mathbf{C}$ is recommended [14][15]. To enhance robustness the intra-subspace projection dominance (IPD) property [14] is used to threshold the representation matrix $\mathbf{C}$:

$$\mathbf{C} \leftarrow \lceil \mathbf{C} \rceil_d \quad (4)$$

where the operator $\lceil \circ \rceil_d$ is applied column-wise, retaining the $d$ largest coefficients (in absolute value) and setting all remaining entries to zero. Here, $d$ denotes the subspace dimension, which is generally unknown *a priori* for a given high-dimensional dataset.

## 3. DERIVATION OF SUBSPACE CLUSTERING MIXTURE MODEL

### 3.1 Taylor expansion of nonlinear mixture model

The NMM used in nonlinear BSS is given by [16]:

$$\mathbf{x}_n = \mathbf{f}(\mathbf{s}_n) \quad n = 1,\ldots,N \quad (5)$$

where, in the BSS terminology, $\mathbf{x}_n$ denotes the observation (mixture) vector and $\mathbf{s}_n \in \mathbb{R}^{C\times 1}$ denotes the corresponding source vector, whose elements are random variables. In this work, however, no statistical assumptions are imposed on the sources. Since our focus is on the subspace clustering interpretation of (5), we assume $C$<$N$. The mapping $\mathbf{f}: \mathbb{R}^C \to \mathbb{R}^D$ is an unknown multivariate nonlinear function satisfying $\mathbf{f}(\mathbf{s}_n) = \left[f_1(\mathbf{s}_n) \ldots f_D(\mathbf{s}_n)\right]^{\mathrm{T}}$ and $\left\{f_i: \mathbb{R}^C \to \mathbb{R}\right\}_{i=1}^{D}$. Furthermore, we assume each component $\{f_i\}_{i=1}^{D}$ is differentiable up to an unknown order $K$.

We next re-write (5) using the $K$th order Taylor expansion around the point $\mathbf{s}_0 = \mathbf{0}_{C\times 1}$ [17]. For notational simplicity, the sample index $n$ is omitted throughout the derivations. Following [17], we express the Taylor expansion of a vector-valued function with a vector argument up to arbitrary order $K$ using tensorial notation, [35].[1] Tensors are denoted by underlined bold capital letters (e.g., $\underline{\mathbf{X}}$). The $k$th order derivative is represented by a tensor of order 1+$k$:

$$\underline{\mathbf{A}}^k \in \mathbb{R}^{D\times \overbrace{C\times\ldots\times C}^{k\ times}} \quad k = 0,\ldots,K \quad (6)$$

whose elements are the corresponding partial derivatives of the order $k$: $\left[\underline{\mathbf{A}}^k\right]_{dc_1\ldots c_k} = \partial^k f_d(\mathbf{s}_0)/\partial s_{c_1}\ldots\partial s_{c_k}$. We further assume that component functions $\{f_i\}_{i=1}^{D}$ possess the same degree of smoothness, i.e., $K_1$=...=$K_D$=$K$. Using the mode-$r$ tensor product, [36], the $k$th term of the Taylor expansion can be written as:

$$\mathbf{x}^k = \frac{1}{k!}\underline{\mathbf{A}}^k \times_1 \mathbf{I} \times_2 \mathbf{s}^{\mathrm{T}} \ldots \times_{k+1} \mathbf{s}^{\mathrm{T}} \quad k = 0,\ldots,K \quad (7)$$

where $\mathbf{I} \in \mathbb{R}^{D\times D}$ denotes the identity matrix, $\mathbf{x}^k \in \mathbb{R}^{D\times 1}$, and (7) is a Tucker tensor model [36] [37]. By unfolding $\underline{\mathbf{A}}^k$ along the mode-*1*, $\mathbf{x}^k$ can be expressed in matrix form as:

[1] The literature primarily considers Taylor expansions involving first-order (Jacobian) and second-order (Hessian) derivatives of vector-valued functions with vector arguments, as well as first-order derivatives of matrix-valued functions with matrix arguments. As noted in [32], third- and higher-order derivatives are required only occasionally. The main obstacle to their use is the rapidly increasing notational complexity. For a vector-valued function with a vector argument, such as (5), the Jacobian is a matrix, whereas the Hessian is a third-order tensor. Consequently, analytical expressions become increasingly cumbersome as higher-order terms are included in the Taylor expansion. References [33] and [34] (Chapters 9 and 10) recommend a differential-based approach for computing first- and second-order derivatives of vector- and matrix-valued functions. However, no analogous methodology is provided for higher-order derivatives.

$$\mathbf{x}^k = \frac{1}{k!}\mathbf{A}_{(1)}^k \left[\underbrace{\mathbf{s}\otimes ...\otimes \mathbf{s}}_{k\text{ times}}\right] k=0,...,K \tag{8}$$

where ⊗ denotes the Kronecker product. Consequently, the data vector in (5) can be re-written as:

$$\mathbf{x} = \sum_{k=0}^{K}\frac{1}{k!}\underline{\mathbf{A}}^k \times_1 \mathbf{I}\times_2 \mathbf{s}^{\mathrm{T}}...\times_{k+1}\mathbf{s}^{\mathrm{T}} = \sum_{k=0}^{K}\frac{1}{k!}\mathbf{A}_{(1)}^k\left[\underbrace{\mathbf{s}\otimes ...\otimes \mathbf{s}}_{k\text{ times}}\right] \tag{9}$$

The elements of the vector $\underbrace{\mathbf{s}\otimes ...\otimes \mathbf{s}}_{k\text{ times}}$ are the monomials of order $k$, i.e.,

$$\underbrace{\mathbf{s}\otimes ...\otimes \mathbf{s}}_{k\text{ times}} = \left[\left\{s_{c_1}^{q_1}...s_{c_p}^{q_p}\right\}_{c_1,...,c_p=1;\,p=1}^{C;\ k}\right] \quad . \tag{10}$$
$$\text{s.t. } q_1,...,q_p \in \{0,1,...,p\}\ \ and\ \sum_{i=1}^{p} q_i = p$$

Therefore, the Taylor expansion in (9) can be written as:

$$\mathbf{x} = \mathbf{f}(\mathbf{s}_0) + \mathbf{A}\overline{\mathbf{s}} \tag{11}$$

where $\mathbf{A}\in\mathbb{R}^{D\times\sum_{k=1}^{K}C^k}$ denotes the block matrix written as:

$$\left[\mathbf{A}_{(1)}^1 \middle| \mathbf{A}_{(1)}^2 \middle| ... \middle| \mathbf{A}_{(1)}^K\right], \tag{12}$$

and symbol "|" separates individual blocks. The vector $\overline{\mathbf{s}}\in\mathbb{R}^{\sum_{k=1}^{K}C^k\times 1}$ is defined as:

$$\overline{\mathbf{s}} = \left[s_1...s_C \left\{s_{c_1}^{q_1}s_{c_2}^{q_2}\right\}_{c_1,c_2=1}^{C} ... \left\{s_{c_1}^{q_1}...s_{c_K}^{q_K}\right\}_{c_1,c_2,...,c_K=1}^{C}\right]^{\mathrm{T}}. \tag{13}$$

Introducing the notation $\overline{\mathbf{x}} = \mathbf{x} - \mathbf{f}(\mathbf{s}_0)$, (11) can be written compactly as:

$$\overline{\mathbf{x}} = \mathbf{A}\overline{\mathbf{s}}\ . \tag{14}$$

To establish connection with SC we assume:

$$\mathbf{s} = s_c\mathbf{e}_c\ \ c\in\{1,...,C\},\ s_c\in\mathbb{R} \tag{15}$$

where $\mathbf{e}_c$ denotes the vector in the standard Euclidean basis in $\mathbb{R}^C$. Under this assumption, (10) reduces to:

$$\underbrace{\mathbf{s}\otimes ...\otimes \mathbf{s}}_{k\text{ times}} = \left[s_1^k\delta(1-c)\underbrace{0...0}_{C^{k-1}-1\text{ times}}\ s_2^k\delta(2-c)\underbrace{0...0}_{C^{k-1}-1\text{ times}}\ ...s_C^k\delta(C-c)\underbrace{0...0}_{C^{k-1}-1\text{ times}}\right]^{\mathrm{T}} \tag{16}$$

where δ(c) denotes the Kronecker's delta. Equation (16) shows that the $k$th order derivative contributes only $k$th order powers of the corresponding source components $\mathbf{s}$. Consequently, only the columns $\left[f_1^{(k)}(\mathbf{s}_0)....f_D^{(k)}(\mathbf{s}_0)\right]^{\mathrm{T}}$ of $\mathbf{A}_{(1)}^k$ contribute to the Taylor expansion (9), and (14) reduces to a linear model with $\mathbf{A}\in\mathbb{R}^{D\times KC}$, and $\overline{\mathbf{s}}\in\mathbb{R}_{0+}^{KC\times 1}$ is given by:

$$\overline{\mathbf{s}} = \left[s_1\delta(1-c)...s_C\delta(C-c)...s_1^K\delta(1-c)...s_C^K\delta(C-c)\right]^{\mathrm{T}} \tag{17}$$

Finally, if the data point $\mathbf{x}$ belongs to cluster $1\le c\le C$, then combining (12) and (17) yields:

$$\overline{\mathbf{x}} = \underbrace{\left[\mathbf{a}_{c;1}...\mathbf{a}_{c;K}\right]}_{\mathbf{A}_c}\underbrace{\left[s_c\cdots s_c^K\right]^{\mathrm{T}}}_{\mathbf{s}_c} \tag{18}$$

where $\mathbf{a}_{c,k} = \left[\partial^k f_1(\mathbf{s}_0)/\partial s_c{}^k ... \partial^k f_D(\mathbf{s}_0)/\partial s_c{}^k\right]^{\mathrm{T}}$, $k$=1,...,$K$.

### 3.2 Subspace dimension and degree of smoothness

A direct comparison of (1) and (18) shows that $\mathbf{A}_c\in\mathbb{R}^{D\times K}$, which corresponds $\mathbf{A}_c\in\mathbb{R}^{D\times d_c}$ in (1), forms a basis for subspace $S_c$, while $\boldsymbol{s}_c$ corresponds to the representation vector $\boldsymbol{z}_c$ in (1). Consequently, the linear SC model (18) derived from the NMM (5) establishes the following relationship between the subspace dimension and the degree of smoothness:

$$\left\{d_c = d = K\right\}_{c=1}^{C} \tag{19}$$

where the equality follows from the assumption $K_1$=...=$K_D$=$K$ introduces in (6). Although the derivation (6) to (9) is valid for an arbitrary Taylor expansion of order $K$, in practice the nonlinear mappings is expected to be only moderately smooth, implying a relatively small value of $K$. Consequently, the low-dimensional subspace assumption underlying the linear SC model in (1) admits a natural physical interpretation. Equation (18) further yields the following matrix representation:

$$\overline{\mathbf{X}} = \mathbf{A}\overline{\mathbf{S}}\text{, such that: } \mathbf{A} = \left[\mathbf{A}_1,...,\mathbf{A}_C\right]\in\mathbb{R}^{D\times KC} \tag{20}$$

where $\left\{\mathbf{A}_c\in\mathbb{R}^{D\times K}\right\}_{c=1}^{C}$ are given with (18), and $\overline{\mathbf{S}}\in\mathbb{R}_{0+}^{KC\times N}$ where each column vector $\left\{\overline{\mathbf{s}}_n\right\}_{n=1}^{N}$ is given with (17).

### 3.3 Number of anchors, degree of smoothness, and sparsity constraint

Reference [11] proposed the LAPIN algorithm for SC based on the dictionary-learning model:

$$\min_{\mathbf{A},\mathbf{Z},\mathbf{F}}\left\|\mathbf{X}-\mathbf{A}\mathbf{Z}\right\|_F^2 + \lambda tr\left(\mathbf{F}^{\mathrm{T}}\mathbf{L}_{\mathbf{G}}\mathbf{F}\right)$$
$$\text{such that: } \mathbf{Z}^{\mathrm{T}}\mathbf{1}=\mathbf{1}, \mathbf{Z}\ge\mathbf{0}, \left\|\mathbf{z}_n\right\|_0 \le l, \mathbf{F}\in\mathbb{R}^{N\times C}, \mathbf{F}^{\mathrm{T}}\mathbf{F}=\mathbf{I} \tag{21}$$

where $\mathbf{A}\in\mathbb{R}^{D\times M}$ denotes the dictionary, $\mathbf{Z}\in\mathbb{R}_{0+}^{M\times N}$ is a nonnegative representation matrix, $M$ is the number of anchors, $l$ is the sparsity level for the representation vector $\mathbf{z}_n$, $\mathbf{L}_\mathbf{G}$ is graph Laplacian of the bipartite graph $\mathbf{G}\in\mathbb{R}_{0+}^{(N+M)\times(N+M)}$ constructed from $\mathbf{Z}$, and $\mathbf{F}$ is the cluster indicator matrix. Comparing the proposed model (20) with (21) yields:

$$M = \sum\nolimits_{c=1}^{C} K = KC \tag{22}$$

$$l=K\text{, i.e., } \left\{\left\|\mathbf{z}_n\right\|_0 = K\right\}_{n=1}^{N}. \tag{23}$$

Thus, the Taylor expansion of order $K$ admits two complementary interpretations: it determines both the number of anchors through (22) and the sparsity level through (23). Comparing (23) with (19) further gives:

$$l=K=d. \tag{24}$$

Equation (24) provides a theoretical interpretation of the sparsity constraint commonly employed in the SC algorithms. An empirical study [11] showed that:

$$0.2\le M/N\le 0.4 \tag{25}$$

provides stable clustering performance. Assuming that the number of samples $N$ and the number of clusters $C$ are known, which is a common assumption in many SC algorithms, the interval in (25) determines a corresponding range for $M$. Equation (22) then yields associated range of $K$, which by virtue of (19), directly translates into an estimate of the subspace dimension $d$. These estimates define the search interval for the parameter $d$ used in the IPD-based post-processing of the representation matrix $\mathbf{C}$ in (4). Since most SC algorithms assume the availability of a validation dataset for hyperparameters tuning, the optimal value of $d$ can be selected using external labels-based clustering metrics.

## 4. EXPERIMENTAL RESULTS

The theoretical relationships established in (19), (22)–(25) are evaluated on six benchmark datasets: EYaleB (2342 samples, 38 subjects), ORL (400, 40), MNIST (10000, 10), USPS (1000, 10), COIL20 (1400, 20), and COIL100 (7220, 100). However, the comparative evaluation in Section 4.1 is performed on 100 randomly generated subsets containing 43, 7, 50, 50, 22, and 22 samples per class for EYaleB, ORL, MNIST, USPS, COIL20, and COIL100, respectively.

### 4.1 Estimation of bounds on dimension of subspaces

We first estimate the admissible range of the subspace dimension $d$ using the theoretical relationship $M=KC$ in (22) together with empirical guidelines for the number of anchors reported in [20]. For Yale B dataset, with $C=15$, $M$ was estimated as $2^3 \leq M \leq 2^7$. By using equations (19) and (22) we obtain $d=K \in [1, 9]$. For ORL dataset $2^4 \leq M \leq 2^8$ for $C=40$ subjects resulting in $d=K \in [1, 7]$. For handwritten dataset and COIL20, [20] $2^6 \leq M \leq 2^{10}$, leading to the initial estimate $d=K \in [6, 100]$. Equation (25) provides tighter bounds. It translates for MNIST and USPS datasets into $100 \leq M \leq 200$ and for COIL20, and COIL100 datasets into $88 \leq M \leq 176$ and $440 \leq M \leq 880$. Based on (22) we obtain for MNIST and USPS $d=K \in [10, 20]$, and for COIL20 and COIL100 $d=K \in [4, 9]$. Combining these predictions with the ones obtained previously we propose for MNIST and USPS $d=K \in [6, 20]$ and for COIL20 and COIL100 $d=K \in [4, 9]$. These predictions agree well with the values reported in the literature: $d \approx 12$ for MNIST and USPS [27], $d \approx 9$ for EYaleB [32] and COIL20 [26]. The only exception is ORL, for which [26] recommends $d \approx 9$, whereas the proposed analysis predicts $d \approx 7$.

To validate the predicted intervals, IPD-based post-processing [14] was applied to the representation matrices produced by GMC-LRSSC [8], SSC [7], LRR [6], and LRR-$S_{1/2}$/LRR-$S_{2/3}$ [12]. For all datasets, the parameter $d$ was cross-validated over the range (1:20). Table 1 reports the value of $d$ yielding the highest clustering accuracy for each algorithm-dataset pair; "-" indicates that IPD post-processing did not improve performance. In all cases, the experimentally selected values fall within the theoretically predicted intervals, including the ORL dataset, thereby supporting the proposed analysis and suggesting that the commonly adopted value $d \approx 9$ for ORL may be overly large.

Table 1. Subspace dimension corresponding with the highest accuracy on EYaleB, ORL, MNIST, USPS, COIL20 and COIL100 datasets.

| | GMC-LRSSC | LRR | SSC | LRR $S_{1/2}$ | LRR $S_{2/3}$ |
|---|---|---|---|---|---|
| EYALEB | 8 | 9 | 3 | 9 | 9 |
| ORL | 5 | 5 | 7 | 6 | 6 |
| MNIST | 12 | 11 | 10 | 9 | 9 |
| USPS | 12 | 12 | 11 | 12 | 14 |
| COIL20 | 6 | 7 | 8 | 7 | 8 |
| COIL100 | - | 9 | 2 | 4 | 4 |

### 4.2 Comparison with the estimated sparsity level

The LAPIN algorithm was empirically shown in [11] to achieve consistently good clustering performance on four benchmark face datasets when the sparsity level satisfies:

$$5 \leq l \leq 20. \tag{26}$$

According to the theoretical relationship established in (24), the sparsity level equals the subspace dimension, i.e., $l=d$. Consequently, the experimentally selected values of $d$ reported in Table 1 can also be interpreted as estimates of the optimal sparsity level. Although, except for EYaleB, the datasets considered here differ from those used in [11], the observed values lie predominantly within the interval (26). This agreement provides additional empirical support for the theoretical relationship $l=d$.

## 5. CONCLUSION

In this paper, we established an equivalence between the nonlinear mixture model on manifolds and the linear union-of-subspaces (UoS) model used in subspace clustering. This result provides a physical interpretation for the commonly assumed low-dimensional structure of the UoS model by relating the subspace dimension to the degree of smoothness of the underlying nonlinear mappings. Furthermore, the proposed analysis establishes explicit relationships among the number of anchors, the degree of smoothness, the subspace dimension, and the sparsity level of the representation matrix. The derived relationships enable principled estimation of the subspace dimension, which can be used to guide the IPD-based post-processing of learned representation matrices. Experimental results on six benchmark datasets demonstrate that the predicted parameter ranges are consistent with those yielding the highest clustering performance across several representative subspace clustering algorithms.

## ACKNOWLEDGMENT

This work was supported by the Croatian Science Foundation grant under the project number HRZZ-IP-2022-10-6403.

## REFERENCES

[1] R. Xu and D. Wunsch, "Survey of clustering algorithms," *IEEE Transactions on Neural Networks*, vol. 16, no. 3, pp. 645-678, 2005.

[2] J. Wight and Y. Ma, High-Dimensional Data Analysis with Low-Dimensional Models - Principles, Computation and Applications, Cambridge University Press, 2022.

[3] L. Wang and C. Pan, "Robust level set image segmentation via a local correntropy-based k-means clustering," *Pattern Recognition*, vol. 47, no. 5, pp. 1917-1925, 2014.

[4] W. Wu, and M. Peng, "A data mining approach combining *k*-means clustering with bagging neural network for short-term wind power forecasting," *IEEE Internet of Things Journal*, vol. 4, no. 4, pp. 979-986, 2017.

[5] M. Udell, and A. Townsend, Why are big data matrices approximately low rank? *SIAM J. Math. Data Sci*., vol. 1, no. 1, pp. 144–160, 2019.

[6] G. Liu, Z. Lin, S. Yan, J. Sun, Y. Yu, and Y. Ma, "Robust recovery of subspace structures by low-rank representation," *IEEE Trans. Patt. Anal. Mach. Intell.*, vol. 35, no. 1, pp. 171–184, 2013.

[7] E. Elhamifar and R. Vidal, "Sparse Subspace Clustering: Algorithm, Theory, and Applications," *IEEE Trans. Patt. Anal. Mach. Intell.*, vol. 35, no. 1, pp. 2765-2781, 2013.

[8] M. Brbić and I. Kopriva, " $\ell_0$ Motivated Low-Rank Sparse Subspace Clustering," *IEEE Trans. Cyber.*, vol. 50, no. 4, pp. 1711-1725, 2020.

[9] C. Lu, J. Feng, Z. Lin, T. Mei, and S. Yan, "Subspace Clustering by Block Diagonal Representation," *IEEE Trans. Patt. Anal. Mach. Intell.*, vol. 41, no. 2, pp. 487-501, 2018.

[10] C.-G. Li and R. Vidal, "A structured sparse plus structured low-rank frame work for subspace clustering and completion," *IEEE Trans. Signal Process.*, vol. 64, no. 24, pp. 6557–6570, Dec. 2016.

[11] F. Nie, W. Chang, R. Wang and X. Li, "Learning an Optimal Bipartite Graph for Subspace Clustering via Constrained Laplacian Rank," *IEEE Transactions on Cybernetics*, vol. 53, No. 2, pp. 1235-1246, 2023.

[12] H. Zhang, J. Yang, F. Shang, C. Gong, and Z. Zhang, "LRR for subspace segmentation via tractable Schatten-*p* norm minimization and factorization," *IEEE Trans. Cyber.*, vol. 49 (5), pp. 1722-1724, 2019.

[13] U. von Luxburg, "A tutorial on spectral clustering," *Stat. Comput.*, vol. 17, no. 4, pp. 395–416, 2007

[14] X. Peng, Z. Yu, Z. Yi and H. Tang, "Constructing the l2-graph for robust subspace learning and subspace clustering," *IEEE Trans. Cyber.*, vol. 47, no. 4, pp. 1053–1062, 2017.

[15] J. Yang, J. Lang, K. Wang, P. L. Rosin and M.-H. Yang, "Subspace Clustering via Good Neighbors," *IEEE Trans. Patt. Anal. Mach. Intell.*, vol. 42, no. 6, pp. 1537-1544, 2020.

[16] C. Jutten, M. Babaie-Zadeh and J. Karhunen, "Nonlinear mixtures," Chapter 14 in *Handbook of Blind Source Separation*, Academic Press, P. Common and C. Jutten Eds., pp. 549-592, 2010.

[17] I. Kopriva, I. Jerić, M. Filipović and L. Brkljačić, "Empirical Kernel Map Approach to Nonlinear Underdetermined Blind Separation of Sparse Nonnegative Dependent Sources: Pure Components Extraction from Nonlinear Mixtures Mass Spectra," *J. Chemometrics*, vol. 28, pp. 704-715, 2014.

[18] C. You, C. Li, D. P. Robinson and R. Vidal, "Self-Representation Based Unsupervised Exemplar Selection in a Union of Subspaces," *IEEE Trans. Patt. Anal. Mach. Intel.*, vol. 44, no. 5, pp. 2698-2711, 2022.

[19] F. Nie, C. Liu, R. Wang, Z. Wang and X. Li, "Fast Fuzzy Clustering Based on Anchor Graph," *IEEE Trans. Fuzzy Syst.*, vol. 30, no. 7, pp. 2375-2387, 2021.

[20] S. Shi, F. Nie, R. Wang and X. Li, "Fast Multi-view Clustering via Prototype Graph," *IEEE Trans. Knowl. Data Eng.*, vol. 35, no. 1, pp. 443-455, 2023.

[21] T. Hastie and P. Y. Simard, "Metrics and models for handwritten character recognition," in *Proc. Stat. Sci.*, 1998, pp. 54–65.

[22] J. Lipor and L. Balzano, "Leveraging union of subspace structure to improve constrained clustering," in *International Conference on Machine Learning (ICML)*, pp. 2310-2139, 2017.

[23] K.-C. Lee, J. Ho and D. Kriegman, "Acquiring Linear Subspaces for Face Recognition under Variable Lighting," *IEEE Trans. Patt. Anal. Mach. Intell.*, vol. 27, no. 5, pp. 684-698, 2005.

[24] A.S. Georghiades, P. N. Belhumeur and D. J. Kriegman, "From few to many: Illumination cone models for face recognition under variable lighting and pose," *IEEE Trans. Pattern Anal. Mach. Intell.,* vol. 23, pp. 643–660, 2001.

[25] F. S. Samaria and A. C. Harter, "Parameterisation of a stochastic model for human face identification," in *Proc. of the 1994 IEEE WACV*, 1994, pp. 138–142.

[26] S. A. Nene, S. K. Nayar and H. Murase, "Columbia object image library (coil-100)," CUCS-006-96, Dept. of Computer Science, Columbia Univ., New York, NY, USA, https://www1.cs.columbia.edu/CAVE/software/softlib/coil-100.php, 1996.

[27] Y. LeCun, L. Bottou, Y. Bengio and P. Haffner, "Gradient-based learning applied to document recognition," *Proc. IEEE, vol. 86*, pp. 2278–2324, 1998.

[28] J. J. Hull, "A database for handwritten text recognition research," *IEEE Trans. Pattern Anal. Mach. Intell.,* vol. 16, pp. 550–554, 1994.

[29] C. You, C. G. Li, D. P. Robinson and R. Vidal, "Is Affine Constraint Needed for Affine Subspace Clustering?" in *Proc. 2019 IEEE ICCV*, 2019, pp. 9915-9924.

[30] I. Selesnick, "Sparse regularization via convex analysis," *IEEE Trans. Signal Process.*, vol. 65, no. 17, pp. 4481–4494, Sep. 2017.

[31] K.-C. Lee, J. Ho and D. Kriegman, Acquiring linear subspaces for face recognition under variable lighting, *IEEE Trans. Pattern Anal. Mach. Intell*., vol. 27, no. 5, pp. 684-98, 2005.

[32] J. R. Magnus, "On the concept of matrix derivative", *J. Multivariate Analysis*, vol. 101, pp. 2200-2206, 2010.

[33] J. R. Magnus and H. Neudecker, "Matrix Differential Calculus with Applications to Simple, Hadamard, and Kronecker Products", *J. Math. Psychol.*, vol. 29, No. 4, pp. 474-492, 1985.

[34] J. R. Magnus and H. Neudecker, Matrix Differential Calculus with Applications in Statistics and Economics, revised edition, John Wiley, 1999.

[35] H. A. L. Kiers, "Towards a standardized notation and terminology in multiway analysis," *J. Chemometrics*, vol. 14, no. 3, pp. 105-122, 2000.

[36] T. G. Kolda and B. W. Bader, "Tensor Decompositions and Applications," *SIAM Review*, vol. 51, no. 3, pp. 455-500, 2009.

[37] L. R. Tucker, "Some mathematical notes on three-mode factor analysis," *Psychometrika*, vol. 31, pp. 279-311, 1966.

[38] M. Abdolali, N. Gillis, and M. Rahmati, "Scalable and robust sparse subspace clustering using randomized clustering and multilayer graphs," *Sig. Proc.*, vol. 163, pp. 166-180, 2019.